# Mitigating severe over-parameterization in deep convolutional neural networks through forced feature abstraction and compression with an entropy-based heuristic

Nidhi Gowdra✉, Roopak Sinha, Stephen MacDonell & Wei Qi Yan
*School of Engineering, Computer and Mathematical Sciences,
Auckland University of Technology, New Zealand*
nidhi.gowdra@aut.ac.nz, roopak.sinha@aut.ac.nz, stephen.macdonell@aut.ac.nz, weiqi.yan@aut.ac.nz

**Abstract**

*Convolutional Neural Networks (CNNs) such as ResNet-50, DenseNet-40 and ResNeXt-56 are severely over-parameterized, necessitating a consequent increase in the computational resources required for model training which scales exponentially for increments in model depth. In this paper, we propose an Entropy-Based Convolutional Layer Estimation (EBCLE) heuristic which is robust and simple, yet effective in resolving the problem of over-parameterization with regards to network depth of CNN model. The EBCLE heuristic employs a priori knowledge of the entropic data distribution of input datasets to determine an upper bound for convolutional network depth, beyond which identity transformations are prevalent offering insignificant contributions for enhancing model performance. Restricting depth redundancies by forcing feature compression and abstraction restricts over-parameterization while decreasing training time by* **24.99% - 78.59%** *without degradation in model performance. We present empirical evidence to emphasize the relative effectiveness of broader, yet shallower models trained using the EBCLE heuristic, which maintains or outperforms baseline classification accuracies of narrower yet deeper models. The EBCLE heuristic is architecturally agnostic and EBCLE based CNN models restrict depth redundancies resulting in enhanced utilization of the available computational resources. The proposed EBCLE heuristic is a compelling technique for researchers to analytically justify their HyperParameter (HP) choices for CNNs. Empirical validation of the EBCLE heuristic in training CNN models was established on five benchmarking datasets (ImageNet32, CIFAR-10/100, STL-10, MNIST) and four network architectures (DenseNet, ResNet, ResNeXt and EfficientNet B0-B2) with appropriate statistical tests employed to infer any conclusive claims presented in this paper.*

## 1. INTRODUCTION

A key challenge in designing CNN models is estimating their appropriate size (depth & breadth) since these parameters are critical in establishing a CNN's representational capacity [1]. Initially, designing a CNN model seems trivial as there exists mathematical proof, that any decision boundary can be approximated with a single sufficiently broad hidden layer [2]. Training a CNN model with a single broad layer is difficult, introducing afflictions like over-fitting [3] or under-fitting, increased susceptibility to spatial variances in the input data [4] and ineffective feature extractions [5].

Training deeper CNN models with stacked hidden layers can help mitigate training afflictions and improve model performance since deeper layers learn more complex feature representations. In the worst-case deeper layers can resolve into identity transformations [7] without incurring any performance penalties. Although training very deep CNNs with up to a thousand layers can be achieved, utilizing current computational hardware, practical limitations such as time and cost for training such very deep CNN models become prohibitively expensive relative to shallower models.

Furthermore, training very deep, yet narrow CNN models present similar training afflictions when compared to a shallow, yet very broad CNN model [6]. Training inefficiencies also become especially apparent when empirically shallower models learn the same functional representations and characteristics as deeper models [8]. Diminishing returns for the ResNet architecture with an exponential increase in layer depth results in marginal gains of accuracy as indicated in [10] where a nominal increase in classification accuracy of 1.1% is achieved from an additional 117 convolutional layers.

While deeper CNNs are extensively employed for computer vision problems like image classification, efficient CNN architectures optimizing CNN depth [9], breadth [10] or both [12] are gaining prominence, with some architectures achieving good results even with limited computing infrastructures [13]. New and emerging research trends are focussing on compression and pruning of very deep CNNs to reduce the associated computational overheads arising in training excessively deep models [14].

Diminishing model performance with exponential increases in the total number of model parameters can be witnessed in all CNN architectures employing skip connections as these shortcut paths essentially produce an ensemble of shallower networks [11]. Therefore, experimental data suggests there might be an upper bound to model depth beyond which there is an insignificant contribution of feature abstraction from the additional layers and could even be detrimental to model performance as these additional layers might induce over-fitting. The absence of a general framework to effectively determine a CNN model's size stems from an incomplete understanding of the underlying mechanisms of action.

The lack of a thorough understanding of the internal workings of CNNs have led to conjecture and opinionated postulations of re- searchers in justifying architecture selections. The rapid progress in the domain of computer vision has also created hurdles for performance evaluations of broader and deeper residual networks. A conclusive determination of optimal CNN HyperParameters (HPs) cannot be ascertained due to the inherent immense complexity and variability involved in computer vision tasks. HP optimization is currently dominated by practitioners knowledge on the subject matter, the computer vision task at hand and available computational resources. Scientific evaluations of model performance with regards to network depth suggests, diminishing returns in model performance for excessive network depths [9] and as such, more investigation is needed to regulate CNN model depth.

Contemporary compression and pruning techniques sacrifice classification accuracy for decreasing model training times. Forcing feature abstraction and compression by constraining model depth to the entropic data distribution of the input dataset should prove be a targeted solution since critical feature information is retained compared to stochastic methods of pruning or compression.

In this paper, we highlight CNN model training inefficiencies in deep CNNs and propose an Entropy-Based Convolutional Layer Estimation (EBCLE) heuristic to eliminate residual depth redundancies improving feature compression. Adequate feature compression enhances hierarchical feature abstraction and reduce model training time. The proposed EBCLE heuristic provides an upper bound value for model depth in CNN architectures based on the a priori knowledge of the entropic data distribution of the input dataset.

A heuristic is justified since it is well understood that optimality in terms of hidden layers cannot be accurately determined for CNN models [7]. Furthermore, using entropy-based approaches for effective feature extraction is well grounded in literature [15]. However, the problem with using entropy measures is that, there are numerous entropy measurements for digital data and it is imperative that a suitable entropy measure is utilized.

Shannon's Entropy (SE) [16] is a measure used primarily in digital communication to improve the latency between information transmission through compression. We hypothesize that, feature compressibility and abstraction in CNNs can only ever meet but not exceed the SE measure, since it is the theoretic limit of digital data compressibility. Thus, a function of SE is justified for estimating the upper bound of convolutional depth in CNNs as these layers are principally involved in feature extraction/information processing.

As CNNs disregard the spatial orientation of the features in an image [17], utilizing SE measures for information measurement is warranted since SE measures are independent of spatial variances in the data. The inherent problem in limiting network depth of CNNs is that, it invariably restricts the information extraction capability i.e. decreases learning capacity of the network since representational power of a CNN is proportional to its size (i.e. depth × breadth).

CNN models using simplistic datasets (MNIST) do not suffer significantly from a decrease in learning capacity due to severe model over-parameterization but, a more pronounced effect can be witnessed for complex natural image datasets such as CIFAR- 100 or ImageNet due to their associated feature complexities. In order to alleviate the decrease in learning capacity from constraining network depth, a subsequent increase in convolutional breadth is necessary as discussed in Section 2. In our experimentation, shallower yet broader CNN models are shown to maintain or even outperform baseline test-set classification accuracies for all the five benchmarking datasets (MNIST, CIFAR-10, CIFAR-100, STL-10 and ImageNet32), while model training time decreased by 45.22% on average across three different CNN architectures (ResNet [7], DenseNet [18] and ResNeXt [19]). Furthermore, our proposed EBCLE heuristic outperforms dynamically scaling approaches utilizing depth and breadth coefficients [12].

The contributions of this paper are as follows:

- We propose an accurate heuristic to determine an upper bound to convolutional depth using Shannon's entropy measure for forced feature compression and abstraction.

- We provide empirical evidence to demonstrate that EBCLE- based shallow neural networks can learn similar high-level feature maps compared to deeper models, as presented in Fig. 3.

- Our proposed entropy-based heuristic reduces CNN model training times by 24.99%-78.59% across three different CNN architectures and five benchmarking datasets without compromising model performance.

- We show that competitive results can be achieved using shallow yet broader CNN models relative to baseline models.

- Our experiments empirically validate and support the findings presented in [9] that, deep CNN models behave as a collection of ensemble networks and the conclusions found in [20] that, wider yet shallower CNN models can learn the same functional representations as deeper yet narrower CNN models with a reduction in the associated trade-off of relative model training time.

The impact of contributions made in this paper are apparent, EBCLE-based CNN models substantially reduce CNN model training time, democratizing research in the domain of computer vision to researchers or practitioners with limited compute capabilities. Accelerated research outputs with the opportunity to test hypotheses rapidly can be achieved through our proposed EBCLE-based heuristic. Furthermore, researchers or practitioners can analytically justify their HyperParameter (HP) choices rather than arbitrarily selecting HP configurations. In general, all Deep Neural Networks (DNNs) exhibiting asymmetries in generalization $\Delta G$ and complexity $\Delta C$ (discussed in Section 2.2) should greatly benefit from feature compression to reduce model training time, regardless of the associated task such as image classification or segmentation.

## 2. BACKGROUND

Consider the task of classifying high-dimensional interpolated data such as images, which can be represented by an under- lying function say, $f(X)$ from a collection of $n$ number of $d$-dimensional input image vectors, $\mathbf{X} = \{x_1, x_2, \cdots, x_n\}$ where $x_i = \langle x^1, \cdots, x^d \rangle | i, d \in \mathbb{Z} > 0$; $f(X) \in \mathbb{R}$. All the images in $\mathbf{X}$ have an associated class label denoted as, $Y = \{y_1, y_2, \cdots, y_n\}$ such that, $(x_i, y_i) \in X \times Y$. The goal of image classification is to learn the underlying functional representation of $\mathbf{X}$ using the $n$ number of input images such that, all images in $\mathbf{X}$ can be linearly separated. Traditional statistical techniques will fail in some classification tasks if the data is represented in high-dimensional vector space and as such alternate methods are needed for accurate classification.

A typical deep CNN model comprises of a convolutional block containing stacked convolutional layers, followed by a pooling layer which is followed by a classification block consisting of multiple fully connected layers with an ultimate softmax activated classification layer. The softmax layer converts all real vector values into class posterior probabilities which sum to 1. Convolutional Neural Networks (CNNs) approximate the underlying functional representation of $\mathbf{X}$ denoted as $\hat{f}(X)$ by projecting the higher-dimensional input vectors of $\mathbf{X}$ into a lower-dimensional vector space say, $\phi(X) = \{\phi(x_1), \phi(x_2), \cdots, \phi(x_n)\}$. A simple regression vector θ at the final classification layer (predominantly softmax activated) can be utilized for linear separation of images in $\mathbf{X}$. The lower-dimensional feature map vector $\vartheta$ produced from $\phi$ are critical for classifying higher-dimensional interpolated data [1]. Functions $f$ and $\hat{f}$ outputs continuous scalar values in the set of rational numbers $\mathbb{R}$ i.e. class posterior probability values provided each input vector in $\mathbf{X}$ has an associated single class label in $Y$.

As output values of $\hat{f}(X)$ are continuous scalar values, they need to be discretized into a single class from all available classes i.e. the number of classes $K$ for the input dataset. Discretization can be achieved using a softmax layer which takes an in- put vector α and outputs a vector of same size β, where $\beta_i = e^{\alpha_i} / \sum_{i=1}^{k} e^{\alpha_i} |\alpha_i; \beta_i \in \alpha; \beta$. The output vector β for the input vector $\alpha$ is normalized such that the logits are in the interval of [0,1] and sum to 1. Consider a binary classification example, lets say for an input image $x_i$ the softmax outputs are ([0.8,0.2]) i.e. an 80% confidence that the input image belongs to class one and a 20% confidence that the image belongs to class two. A threshold function can encode the logits into a one-hot vector such that, $y_i$ = [1,0]. Model performance is dependent on an accurate dimensional reduction of the input $\mathbf{X}$ denoted as $\phi(\mathbf{X})$ as this is challenging for traditional statistical or function mapping techniques, alternatives need to be explored.

Computing $\phi(\mathbf{X})$ could be achieved through feature extraction rather than mapping the underlying function of $\mathbf{X}$. CNNs are state-of-the-art for feature extraction as they utilize convolutional kernels/channels/units/filters which are scanned across the input image in $\mathbf{X}$ producing feature maps. The number of convolutional kernels/channels/units/filters can be denoted as $\chi' | \chi' \in \mathbb{Z}_{>0}$. The $\chi'$ number of convolutional kernels have an associated weight vector $\mathbf{W} = \{\omega_1, \cdots \omega_{\chi'} | \mathbf{W} \in \mathbb{R}\}$ extract lower-dimensional feature information for the input vectors in $\mathbf{X}$ to produce a feature map vector $\phi(\mathbf{X}) = \langle \sum_{i=1}^{n} \phi_j(X_i), \cdots, \phi_{\chi'}(X_i) \rangle$, illustrated in Fig. 2. Closer approximations to $f(X)$ can be accomplished by adjusting the weight vector $\mathbf{W}$ of the $\chi'$ number of conv. kernels until $\hat{f}(X)$ can approximate a one-dimensional projection of $f(X)$. A full-linear separation of $f(X)$ can be achieved for the regression vector $\vartheta$ given an optimal weight selection for $\mathbf{W}$ where, $|\phi| = d'$ is much larger than the $d$-dimensional input vector $|\mathbf{X}| = d$ i.e. $1 \leq d \ll d'$.

Assuming a constant $f(X)$ for $X$, computing $\hat{f}(X)$ is given by Eq. 1 [5],

$$\hat{f}(X) = \langle \phi(\mathbf{X}), \mathbf{W} \rangle = \sum_{j=1}^{\chi'} \phi(\mathbf{X}), \omega_j. \quad (1)$$

$\mathbf{W}$ is the weight vector for which the regression vector $\vartheta$ is optimized utilizing the $n$ number of training images in $\mathbf{X}$ and $\phi(\mathbf{X})$ is the feature map vector computed using the feature vector $\phi(\mathbf{X})$ for an input vector $x_i \in \mathbf{X}$. Theoretically, deeper CNNs have an increased capacity to compute a much closer approximation to $f(X)$ compared to shallower CNNs, since deeper networks can abstract more complex feature maps. This is the reason why the enumeration of the feature map vector $\phi(X)$ exponentially grows for a given underlying function $f(X)$ for a given dataset. A larger feature map vector $\phi(X)$ can also be achieved using a shallower yet broader model since they are functionally equivalent to a deeper yet narrower model in terms of generating feature maps shown, discussed later in this paper and illustrated in Fig. 3.

It is worth emphasizing that the dimensionality of feature map vector $|\phi| = d'$ can only be reduced by increasing the convolutional depth, validating the arguments made for using ever deeper CNN models [7]. An oversight to this argument is that, reductions in dimensionality increasingly deviate the computed $\hat{f}(X)$ from the ideal functional representation since activation functions apply approximations at each layer and residual models behave as an ensemble of smaller networks [9] thus increasing redundant feature extractions.

In Section 3, we mathematically deduce that increasing convolutional depth beyond Shannon's entropy measure adds to redundancies, which is consistent with the findings presented in [9]. Furthermore, in Eq. 1, the problems of using deeper networks become especially apparent as $\chi'$ plays a parallel role to $\phi(\mathbf{X})$, which conforms to the conclusions presented in the previous work of [9] and is the reason why wider residual networks [20] perform better than a thousand layer deep network. The dimension of the feature map vector, $|\phi(X)| = d'$ is often why CNN models over-fit or under-fit to the input data.

## 2.1. Convolutional neural network architectures

### 2.1.1. Residual network (ResNet)

Residual Network (ResNet) [7] using skip connections was proposed to solve the degradation problem existing in deep CNNs. The ResNet architecture is a deployment of residual learning-oriented blocks. The goal of using these blocks is to integrate nonlinear convolutions into residual operations, which greatly reduce the difficulty of linear separation in pattern classifications.

Research into shortcut paths to address the problems of vanishing/exploding gradients has been undertaken since the days of Multi-Layer Perceptrons (MLPs) but, offered little improvements. Variations on the shortcut connection paths have since been used in state-of-the-art CNN models [21] such as ResNets with skip connections between hidden layers to allow for retention of the initial features. ResNet follows the function of error minimization for the input training data **X** given in Eq. 2

$$\hat{f}(X) = \langle \phi(\mathbf{X}) + \mathbf{X}, \mathbf{W} \rangle \qquad (2)$$

### 2.1.2. Densely connected convolutional network (DenseNet)

Densely connected convolutional network (DenseNet) [18], was conceived from a simple idea that the output of any hidden layer $h_i | h_i \in H; i \geq 2$ where, $H$ is the depth, should include the con- catenation of all preceding feature maps produced from $h_i - 1$ layers. DenseNet also obeys the error minimization function given in Eq. 3, which requires the computation of a lower dimensional feature vector from the input data **X**.

$$\hat{f}(\mathbf{X}) = \langle \phi([x_i(1), \dots, x_n(h_i - 1)]), \mathbf{W} \rangle \qquad (3)$$

Growth rate is a critical parameter in DenseNet, viewed as the total amount of feature information contributed by an individual layer to the entire network. DenseNets with similar capacity can have varying classification performance by adjusting the growth rate HyperParameter.

### 2.1.3. Aggregated residual transformations (ResNeXt)

A neuron can be thought of as an aggregation of signal transformations from all the input data. The principle of ResNeXt [19] is to replace the computation of $\varphi(\mathbf{X})$ with the transformation of input **X** as $\tau(\mathbf{X})$. ResNeXt, i.e., aggregated residual transformations, can be represented by the error minimization function given in Eq. 4.

$$\hat{f}(\mathbf{X}) = \langle \sum_{i=1}^{L} \tau(\mathbf{X}_n^L), \mathbf{W} \rangle, \qquad n, L \in \mathbb{Z}_{>0} \qquad (4)$$

Cardinality $L$ is the fixed size of aggregated transformations. Cardinality is an important HyperParameter affecting model capacity similar to network depth.

## 2.2. CNN Optimization

### 2.2.1. Parameter compression and pruning

Compressing CNN models either through spatial or channel de- composition [22] is extensively adopted in practice to increase training efficiency by removing depth redundancies. While channel [23] and spatial [24] pruning show significant reduction in model training time, they inevitably offer lower classification performance compared to a deeper un-pruned CNN models. In Section 2.2.3, the ineffectiveness of spatial compression is discussed. In Section 3, the variances in convolutional outputs from channel pruning are highlighted. CNN parameter pruning is also challenged, as broader models with greater numbers of trainable parameters outperform narrower yet deeper models with lower numbers of trainable parameters in terms of training time as they can be more efficiently computed in parallel.

### 2.2.2. Efficient convolutional neural network (EfficientNet)

The premise behind EfficientNet is that, CNN models are developed with a fixed resource budget and are then scaled up to improve model performance. A uniform compound co-efficient is introduced as an alternative to single-dimension scaling [12]. The authors argue that compound scaling is warranted for increasing the receptive field important to capture fine-grained patterns in large images. In this paper, in Section 5, we present empirical evidence to support targeted depth scaling (utilizing Shannon's entropy measure) and manual width scaling constrained only by computational resource budgets offers similar or even enhanced model performance compared to uniform scaling approaches while significantly decreasing training time.

### 2.2.3. Information theory and entropy

Information theory has wide-ranging applications in interdisciplinary domains such as communication systems and complexity theory. Information theory is a derivative of probability theory where the probability measures of particular events are used to determine the complexity of information contained in events [25]. The equation to determine the total amount of information contained in an input $X$ for a given event $E$ is presented in Eq. 5.

$$I(\mathbf{X}) = ln(1/p_E) = -ln(p_E) \qquad (5)$$

Where, $I(\mathbf{X})$ is the total amount of information contained in the event for the input dimensional vector $\mathbf{X}$. The number of states or independent symbols that a single element for a single instance of $X$, denoted as $\mathbf{x}_i^j$, where $i, j \in \mathbb{Z}_{>0}$ can exist in is denoted by $a$ for an event $\varepsilon$ with the natural log $ln(\cdot)$ representing the probability $p$. As a natural log is used, the unit of measurement is NAT-ural units (nats). A natural log with base $e = 2.7182...$ is appropriate in this context as selecting any other logarithm base would restrict the true measurement of representational power. According to [16], the total amount of information contained in any given data is expressed through its entropy ($E$), which can be calculated using Eq. 6.

$$E(\mathbf{X}) = -\sum_{1}^{n}\sum_{i=0}^{a} p_\epsilon \ln p_\epsilon = \sum_{1}^{n}\sum_{i=0}^{a} p_\epsilon I(x_i), n, a \in \mathbb{Z}_{>0} \quad (6)$$

In Eq. 6, $E$ is the Shannon's Entropy (SE) measure in NATu-ral units (nats), $p_\epsilon$ the probability of choice for $a$ distinct independent symbols and $n$ is the number of training data/images. Eq. 6 also implies that the entropy measure is dependent on the total amount of information in an event and the probability of its stochastic source. In other words, if new events yield no new information, the entropy would be zero. In digital images, the $a$ value for a grayscale image would be 256 for 8-bit images, i.e. $2^8 = 256$ or 0 to 255 distinct gray values. $p_\epsilon$ is the probability of a pixel possessing the gray value $a$.

As determining probability and relative probability measures for digital images is impractical due to the high-dimensional interpolated nature of images, histograms/frequency of pixel intensities are computed instead to calculate close approximations to actual probabilities utilizing the open-source scikit-image library written in python. The same process can be applied for color images but probability measures are computed for every color channel i.e. Red(R), Green(G) and Blue(B) color channels.

It is worth highlighting that images with different spatial configurations have the same entropy measures. The loss of account- ability in measuring spatial configurations is a drawback of CNNs in general [17]. SE calculations also disregard spatial variations during measurement and as such, the SE measure is a perfect metric for quantifying the amount of information $I(\mathbf{X})$ in an image.

Using the skikit-image library, we calculate the entropy measures for all the training images present in the MNIST, CIFAR- 10/100, STL-10 and ImageNet32 datasets, described further in Section 4.1. The SE measures of all the training images are then averaged (as the CNN should be able to generalize between all classes of images) across the entire training set and rounded to two digits. The averaged entropy measures are **MNIST: 2.14, CIFAR- 10 and STL-10: 5.03, CIFAR-100: 4.97** and **ImageNet32: 4.97**. As most of the natural image datasets contain images from much of the same classes, it is not surprising that they have similar entropy measures.

# 3. ENTROPY-BASED LAYER ESTIMATION

There are multiple methods proposed to estimate layer/neural configurations of networks as discussed in Sections 3.0.1 and 3.0.2. In this paper, we primarily focus on feature extraction, abstraction and compression to determine the upper and lower bounds for the input vector and a heuristic upper bound to estimate the number of hidden layers required in a CNN. As computation of $\phi(\mathbf{X})$ is predicated upon the information extracted within the hidden layers of a CNN, discussions around information theory and its principles are warranted and most appropriate.

### 3.1. Mutual information neural estimation (MINE)

Authors in [26] empirically demonstrate that Shannon's entropy-based measures to determine mutual information of images $(\mathbf{x}_i, y_i) \in X \times Y$ decreases the uncertainty in approximating the underlying function $f(\mathbf{X})$ given the computation of conditional entropy. The equation to determine mutual information between two vectors $\mathbf{X}$ and $Z$ is given in Eq. 7,

$$I(\mathbf{X}; Z) \coloneqq E(\mathbf{X}) - E(\mathbf{X}|Z) \quad (7)$$

Where, $E$ is Shannon's entropy measure and $E(\mathbf{X}|Z)$ is the conditional entropy of $\mathbf{X}$ given $Z$. Theoretic proofs of MINE exhibit strong consistency for multi-variate information estimation while capturing non-linear dependencies. Furthermore, MINE has been empirically validated to outperform non-parametric estimation in [27]. MINE performs well for adversarial networks and proves tractable for applications utilizing the **principle of Information Bottleneck (IB)** but, no evidence is presented in terms of its ap- plications in Deep Neural Networks (DNNs). Furthermore, MINE is used as an objective function in adversarial setting to maximize $I(\mathbf{X};Z)$. IB has shown to approximate optimal representations of $\mathbf{X}$ with respect to $Y$ in a discrete setting and with the addition of a small noise in a stochastic setting for both adversarial net- works and DNNs [28]. Therefore, MINE's application is limited for DNNs but offers strong empirical evidence that SE can be utilized as a quantitative metric for information compression in neural net- works outperforming other estimation methods.

### 3.2. Mutual information of layers in deep neural networks

According to the authors in [28], the commonly used Stochastic Gradient Descent (SGD) optimizer in DNNs behaves in two different and distinct phases, Empirical erroR Minimization (ERM) and representation compression, with the phases characterized by variations in the gradients Signal to Noise (SNR) ratios of individual layers. The ERM phase results in a rapid increase of the mutual information $I(X;Y)$ with respect to the class label $Y$ and the com- pression phase (the majority of model training is utilized in this phase) is marked by a slow compression of the feature representation of $\mathbf{X}$. Furthermore, the authors in [28] empirically demonstrate that the optimized layers approach the optimal IB bound which plays a pivotal role for computational and accuracy trade-offs.

In a multi-class classification problem a single-objective optimization $T_\epsilon$ of the hidden network layers ($H'$) between $1 \leq \epsilon \leq H'$ is dependent on a multi-objective optimization of $I_X = I(X; T_\epsilon)$ and $I_Y = I(T_\epsilon; Y)$. The ERM phase of model training minimizes the cross-entropy loss characterized by $I_Y$ while the compression phase optimizes $I_X$ which can be represented as $I_X = E(X) - E(X|T_\epsilon)$, If the input entropy $E(\mathbf{X})$ is invariant, optimizing $E(X|T_\epsilon)$ is sufficient, also known as stochastic relaxation. As CNNs are fundamentally differentiated by their convolutional operations to extract feature representations from input **X**, authors in [28] assert the **entropy growth $\Delta E$ for convolutions is logarithmic** in the number of time steps i.e. $\Delta E \; \alpha \; log(\mathcal{D}_t)$ where, $\mathcal{D}$ is the underlying data distribution from which the independent input samples **X** are obtained.

*There is an exponential decrease in model training time with reduced network depths due to stochastic relaxation.* In other words, the IB bound is greatly responsible in optimizing $I(X; T_\epsilon)$ and since shallower networks have fewer number of hidden layers the representation of $I(\mathbf{X})$ is subsequently constrained and thus will train faster given identical computational resources relative to deeper networks. The decrease in representational capacity and methods of mitigating representational losses for shallower networks are explored in Section 3.1.

A search on the information plane (illustrated as Fig. 1) i.e. $I_\mathbf{X}$ and $I_Y$ could yield an upper bound for convolutional layer estimation but, this method involves a pre-training step which is both computationally and time sensitive. In other words, a trial-and-error approach could be adopted to get an ideal curve for $\hat{Y}$ to minimize $\Delta C$ and $\Delta G$ by evaluating $R$ and varying $D_{IB}$ but, this is time consuming and requires additional computational resources.

We propose a logical upper bound and heuristic convolutional depth in Section 3.3 using only the a priori knowledge of the SE measure of **X** i.e. $E(\mathbf{X})$ without pre-training.

### 3.3. Entropy and convolutional depth

As discussed in Section 3.0.2, authors in [28] assert that exponential decreases in model training times are achieved with a reduction in network depth since the majority of model training time is dedicated to feature compression. However, as discussed in Section 2.2.1, compression or pruning inevitably results in adverse model performance due to the associated loss in model learning capacity. The characteristic nature of deep Convolutional Neural Networks (CNNs) using skip connections (such as the ResNet, DenseNet and ResNeXt architectures discussed in Section 2.1) resolve into an ensemble of shallower networks [11] suggesting limiting convolutional depth to enhance feature compression could potentially decrease training time without a significant impact on model performance.

Limiting convolutional depth will invariably constrain the formation of ensembles of shallower networks and a corresponding expansion of the convolutional breadth (i.e. the number of convolutional kernels/channels/filters/units in a hidden layer) should counteract the problem of decreased model learning capacity. While limiting convolutional depth is in stark contrast to the work done by authors in [24] proposing convolutional channel pruning, there is empirical evidence [8] supporting the fact that shallower networks can learn similar complex feature representations as deep networks, primarily because majority of model training is dedicated to feature compression [28] during which redundant information is compressed and only the most important features improving model performance are retained.

An ideal determination of network depth is impractical due to the fact that residual connections propagate information non-sequentially between layers and they can always learn identity transformations allowing for training of very deep CNNs with up to and beyond a thousand layers [7]. Heuristic optimization of the network depth is desirable since lowering architectural complexity decreases the generalization gap but increases the informational complexity gap, as illustrated in Fig. 1 i.e. allows for a broader representation of **X** in the information plane, $I_\mathbf{X}$ and $I_Y$. Ideally, the level of abstraction within a CNN should be equal to the informational complexity of the input dataset. Although this would be ideal, there are no methods of estimating when this level of abstraction is achieved.

The abstraction capability of CNNs is reliant on the representational power of the model, which refers to the ability of the network to accurately extract and represent information in feature maps. Increasing the representational capacity in CNNs acts as a compensation mechanism for the loss of spatial information during the abstraction process. The representational capacity of CNN models discussed in Section 2.1 is increased with each additional convolutional layer. Note that however, although each additional convolutional layer increases the representational capacity of a CNN, these additional layers might be performing identity transformations which do not contribute in enhancing model performance. Furthermore, the compression phase of model training for deep CNN models require an exponential increase in computational resources and training time.

### 3.3.1. Input compression bound

Authors in [28] proposed a new input compression bound presented as Eq. 9, to replace the generalization bounds defined by classic learning theory presented as Eq. 8.

$$\epsilon^2 < \frac{log|\mathcal{H}_\epsilon| + log 1/\delta}{2n} \qquad (8)$$

Where, $\epsilon$ is the difference in errors between training $\Delta C$ and generalization $\Delta G$ as illustrated in Fig. 1. $\mathcal{H}_\epsilon$ is the $\epsilon$-cover for a depth hypothesis assuming the size $|\mathcal{H}_\epsilon| \sim (1/\epsilon)^d$. $d$, the dimensionality of $n$ number of input samples in **X**. $\delta$, the confidence interval of $\hat{Y}$ is between [0,1].

$$|\mathcal{H}_\epsilon| \sim 2^{|\mathbf{X}|} \rightarrow 2^{T_\epsilon} \qquad (9)$$

Where, the size of input vector **X** is $E(\mathbf{X})$ given **X** is large. $T_\epsilon$ is the single-objective optimization as an $\epsilon$-partition of the input vector **X** of size $2^{E(X|T_\epsilon)}$, assuming $2^{T_\epsilon}$ is the cardinality for a depth hypothesis $H_\epsilon | T_\epsilon \in 1 \leq \epsilon \leq H'$. Furthermore, $|T_\epsilon| \sim \frac{2^{E(\mathbf{X})}}{2^{E(X|T_\epsilon)}} = 2^{I(T_\epsilon;x)}$, discussed in Section 3.0.2. Simplifying Eq. 9, the input compression bound can be presented as Eq. 10,

$$\epsilon^2 < \frac{2^{I(T_\epsilon;x)} + log 1/\delta}{2n} \qquad (10)$$

Assuming an absolute confidence and a finite number of samples images (in-bound disregarding out-of-bound distortions), the compression bound in Eq. 9 is dependent on $I(T_\epsilon;x)$, therefore maximizing $I(T_\epsilon;x)$ should be sufficient for enhanced feature compression.

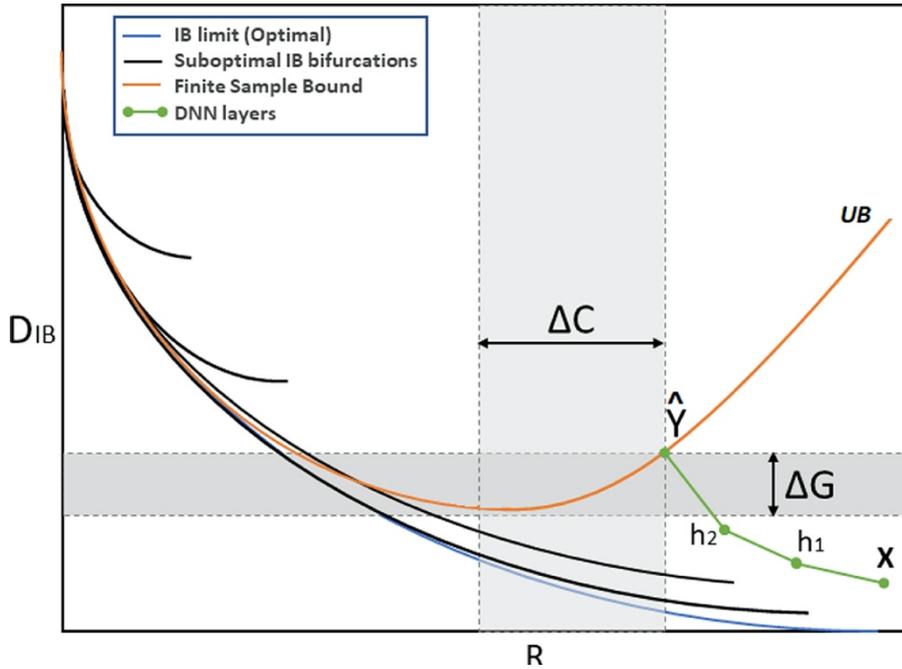

**Fig. 1.** Information plane with a hypothesized layer path in a DNN for finite set of samples in **X**. *C* is the complexity gap and *G* is the generalization gap, *DIB* is the optimal achievable IB limit for samples in **X**, R = I(X;X^) and UB is the upper bound on the out-of-sample IB distortion. Figure reproduced from [29].

**Stochastic relaxation:** As discussed in Section 3.0.2, Layer com- pression can be computed as $\Delta E_i = I(\boldsymbol{X}; T_i) - I(\boldsymbol{X}; T_{i-1})$ for a given hidden layer $h_i \in H'$. Implying an exponential decrease in training time for decrements in the number of hidden layers. Our hypothesis is that redundant information in the input vector **X** can be compressed as $X_E$ for which the resultant feature maps generated by the hidden layers of a CNN cannot exceed Shannon's Entropy measure $E(\boldsymbol{X})$. In other words, limiting convolutional depth ($H'$) with a corresponding increase in convolutional breadth ($\chi'$) for a CNN should exponentially decrease training time without compromising model performance. The progressive increase in spatial convolutions is exponential for $H' \times \chi'$ and is in the order of $2^{H'\chi'}$ [5] i.e. $2^{H'\chi'} \leq I(\boldsymbol{X}_E) \leq E(\boldsymbol{X})$, the CNN compressed feature vector cannot exceed the theoretical limit compression of the input vector. Furthermore, the representational capacity of a CNN is proportional to its size (depth × breadth) and its size is $2^{T_\epsilon}$ (from Eq. 9) i.e. $I(\boldsymbol{X}_E; H') \leq 2^{E(\boldsymbol{X})}$, the information contained in a $H'$ deep CNN is distributed among all of the convolutional kernels which is its representational capacity obtained after the compression phase of model training.

### 3.3.2. Upper bound of convolutional depth

Determining an adequate convolutional depth for which the model provides sufficient dimensionality reductions without introducing inefficiencies or redundancies is a challenging problem. Assume the *d*-dimensional input vector **X** has no redundancies i.e. stochastic noise, in this instance there are no practical ways to apply stochastic relaxation without compromising model performance. This can be considered the lower-limit for hidden layer compression where a convolutional depth estimation is impossible since additional layers will increase model performance significantly.

As most information captured in the real-world has some redundancy, the *n* samples in input **X** can be compressed up to the theoretical limit i.e. Shannon's Entropy (SE) measure *E* (computed using Eq. 6). Lets denote the compressed input vector as $\boldsymbol{X}_E | \boldsymbol{X}_E \leq \boldsymbol{X}$. *Increasing the number of input samples will in effect reduce the suboptimal IB bifurcations as illustrated in Fig. 1.* As discussed in Section 3.0.2, in instances of stochastic relaxation, optimizing $I_{\boldsymbol{X}}$, specifically $E(\boldsymbol{X}|T_\epsilon)$ is adequate for exponential decreases in model training times. We know that entropy growth $\Delta E$ is logarithmic, particularly *ln* (From Eqs. 5 and 6) and $\Delta E_i = I(\boldsymbol{X}; T_i) - I(\boldsymbol{X}; T_{i-1})$ (From Section 3.1.1).

The final convolutional layer output for a CNN model of depth $H'$ is dependent on the previous layer $H' - 1$ and the output for layer $H' - 1$ is determined by the output from its previous $H' - 2$ layer and so on until the first input layer. Therefore, entropy growth $\Delta E$ can be rewritten for the entire convolutional depth of a CNN as Eq.11,

$$\Delta E = ln\left(\sum_{i=2}^{H'} I(\boldsymbol{X}; h_i) - I(\boldsymbol{X}; h_{i-1})\right) \quad (11)$$

A unique characteristic of information propagation in the hidden layers is that $I(\boldsymbol{X}; h_i) \leq I(\boldsymbol{X}; h_{i-1}) \leq I(\boldsymbol{X}_E) \leq I(\boldsymbol{X})$. In other words, any information lost in the initial layer/s cannot be recovered in deeper layers [30]. Furthermore, for any $i \geq j$, $I(Y; \boldsymbol{X}) \geq I(Y; \boldsymbol{X}_E) \geq I(Y; h_j) \geq I(Y; h_i) \geq I(Y; \hat{Y})$ holds true. $I(Y; \hat{Y})$ quantifies the predictive features in **X** for $Y$, determining $I(\boldsymbol{X}; H')$ i.e. the final convolutional layer should yield an upper bound for depth estimation.

The feature map outputs of any convolutional layer is governed by the non-linear activation function $\rho(\cdot)$, most commonly the Rectified Linear Unit (ReLU), $\rho(Z) = max(0, Z)$ for some vector input $Z$ [31]. The activation function essentially bottlenecks information propagation within the hidden layers, such that, $E(\mathbf{X}) \geq E(\mathbf{X}_E)$ and $I(\hat{Y}; \mathbf{X}_E) \leq \rho(I(\mathbf{X}_E))$. The final layer output for a convolutional depth $H'$ requires as an input the compressed vector $\mathbf{X}_E$ (because only the first convolutional layer can accept the uncompressed input vector $\mathbf{X}$, all other layers have the feature map output from the first layer as an input) and is constrained by the activation function i.e. $\rho(I(\mathbf{X}_E; H'))$.

Eq. 11 can be rewritten and reduced as Eq. 12,

$$\Delta E = ln\left(\rho(I(\mathbf{X}_E; H'))\right) \quad (12)$$

$I(\mathbf{X}_E) = 2^{H'\chi'}$ and $I(\mathbf{X}_E; H') \leq 2^{E(\mathbf{X})}$ (from Section 3.1.1 applying stochastic relaxation). The activation function ensures to maximize $I(\mathbf{X}_E; H')$ and equality is achieved if and only if $\hat{\mathbf{X}} = \mathbf{X}$. Therefore, the relationship is invariant for $I(\mathbf{X}_E; H')$ i.e. $\rho(I(\mathbf{X}_E; H')) \leq 2^{E(\mathbf{X})}$. In other words, the final compressed feature vector cannot exceed the theoretical compressibility of the input vector $\mathbf{X}$ and equality is achieved in the best case when $\hat{\mathbf{X}} = \mathbf{X}$. Therefore, Eq. 12 can be rewritten as an upper bound of convolutional depth (which is the Entropy-Based Convolutional Layer Estimation or EBCLE equation) as Eq. 13

$$\Delta E = ln(2^{E(\mathbf{X})}) \quad (13)$$

Where, $\Delta E \in \mathbb{R}_1^+$ is the objective function for maximizing feature compression within the hidden layers of a CNN, given the Shannon's Entropy measure $E(\mathbf{X})$ for an input dataset $\mathbf{X}$. The complexity of $\mathbf{X}$ determined by its entropy measure $E$, indicates higher degree of data complexity requires CNN models with corresponding complexity for dimensionality reduction and linear separation.

Note that since the *EBCLE* heuristic for feature compression belongs to $\mathbb{R}_1^+$, upper and lower bound values are mandatory. It is safe to assume that the upper bound should be used, as using the lower bound might lead to premature feature complexity growth. Utilizing Eq. 13, we can determine the upper bound heuristic for the selected input datasets **CIFAR-10/100, STL-10 and ImageNet32** as **4** and the **MNIST** dataset as **2**.

*3.3.3. Using EBCLE for CNN architectures*
The EBCLE heuristic or $\Delta E$ offers a way to maximize feature compression by utilizing minimal number of hidden convolutional layers and as such this upper bound measure behaves differently for various static CNN architectures due to their architectural constraints. All the CNN architectures employed in this paper have residual learning blocks with stacked convolutional layers, these stacked convolutional layers should not degrade model performance since they can always perform identity transformations [7]. Since stacked convolutions reduce dimensions exponentially, shortcut paths are introduced after each learning block to ensure model performance (feature learning capacity) is not impeded. Therefore, an architectural design lower-bound is placed on model depth.

The design limitations proposed by the authors are, ResNet v1: Depth $= N \times 6 + 2$ [7]; DenseNet: Depth $= N \times 3 + 4$ [18] and ResNeXt: Depth $= N \times 9 + 2$ [19]. $N|N_{>1+}$ is the EBCLE value derived earlier in Section 3.2, i.e. 4 for CIFAR-10/100, STL-10, ImageNet32 and 2 for MNIST. As such, the lower bound for model depth that can be employed for these architectures are, ResNet v1: 8, DenseNet: 7 and ResNeXt: 11.

# 4. EXPERIMENTAL DESIGN

To validate EBCLE as a heuristic, we employ a quantitative approach using five well-known benchmarking datasets, MNIST [32], CIFAR-10/100 [33], STL-10 [34] and ImageNet32 [35]. The selected comparison criteria are, test-set classification accuracy and the model training time. We test the efficacy of EBCLE against deeper ResNet-50, ResNeXt-56 and DenseNet-28 models (deeper, broader models could not be evaluated due to memory constraints), while keeping other HPs such as learning rate and batch size constant with no data excluded or pre-processing steps applied to images in the datasets for three independent evaluation runs. Learning rate and batch size were selected based on configurations by the original authors of the proposed architectural models.

The current consensus is that, using a trial and error methodology to vary HP configurations and using expert domain knowledge, fine-tuning of deep CNN models yield enhanced model performance [7]. Our primary objective for this study is to investigate the relative classification performance of deeper yet narrower and shallower yet broader CNN models with an emphasis placed on training time.

*4.1. Datasets*
The MNIST dataset includes 28 × 28 pixel resolution black and white handwritten digits. MNIST consists of 60,000 training and 10,000 test images split equally into ten classes for each numeric digit. The CIFAR-10/100 datasets includes 50,000 training and 10,000 testing natural color images with a 32 × 32 pixel resolution, split equally into ten/hundred classes for CIFAR-10/100 respectively, which include pictures of airplanes, automobiles, birds and other such natural image classes. The STL-10 dataset includes 500 training and 800 test natural color images split into much of the same classes of natural images but in a higher 96 × 96 pixel resolution, derived from the ImageNet dataset [36]. The Ima- geNet32 dataset is a downsampled (32 × 32 pixel resolution) version of the original ImageNet dataset [36], consisting of a thousand natural image classes.

*4.2. Experimental setup*
The first set of experiments presented in Table 1 were conducted using a single Nvidia 2080ti GPU with an AMD Thread- ripper 1920x CPU and 32GB of RAM, generously provided by InfuseAI Limited (New Zealand). The second set of experiments presented in Table 2 and evaluation of the transfer learning performance were conducted using a single 3070 GPU with a AMD Ryzen R5 2600 CPU with 64GB of RAM, yet again generously provided by InfuseAI Limited (New Zealand). The training-validation split for every model was kept constant at 80%-20% across all datasets. There were no modifications made to the CNN architectures and to ensure reproducibility, no image augmentation techniques were used. HP configuration included using the Adam optimizer with a batch size of 128 and a constant learning rate of 0.001 for 100 epochs. Where possible, official Github repositories were cloned (to preserve anonymity, appropriate acknowledgments/credits will be included in the camera-ready paper) for the four CNN architectures built on the target software platform (Keras with a tensor- flow backend) utilized in this paper.

Table 1. Table of results comparing different CNN models on various benchmarking datasets.

| CNN Model | $H'$ | $\chi'$ | Params. | Acc. (%) | Time (h:m:s) | REL. $\Delta$ (%) |
|---|---|---|---|---|---|---|
| **MNIST dataset** | | | | | | |
| ResNet-50 | 50 | 16 | 760,266 | **99.44** | 0:43:31 | −65.68 |
| **ResNet-EBCLE** | 14 | 24 | 400,474 | 99.42 | 0:14:56 | |
| DenseNet-40 | 40 | 12 | 1,058,866 | **99.60** | 1:03:46 | −78.59 |
| **DenseNet-EBCLE** | 10 | 20 | 210,050 | 99.54 | 0:13:39 | |
| ResNeXt-56 | 56 | 16 | 11,003,712 | 99.15 | 3:48:01 | −63.12 |
| **ResNeXt-EBCLE** | 20 | 16 | 3,893,056 | **99.21** | 1:24:05 | |
| **CIFAR-10 dataset** | | | | | | |
| ResNet-50 | 50 | 16 | 765,098 | 79.85 | 0:38:33 | −36.27 |
| **ResNet-EBCLE** | 26 | 24 | 830,698 | **80.85** | 0:24:34 | |
| DenseNet-40 | 40 | 12 | 1,059,298 | **87.22** | 1:05:38 | −62.09 |
| **DenseNet-EBCLE** | 16 | 20 | 692,690 | 86.35 | 0:24:53 | |
| ResNeXt-56 | 56 | 16 | 11,004,864 | 87.08 | 3:38:47 | −30.08 |
| **ResNeXt-EBCLE** | 38 | 16 | 7,449,536 | **87.29** | 2:32:59 | |
| **STL-10 dataset** | | | | | | |
| ResNet-50 | 50 | 16 | 765,386 | 56.56 | 0:16:35 | −32.73 |
| **ResNet-EBCLE** | 26 | 24 | 838,378 | **60.17** | 0:12:30 | |
| DenseNet-40 | 40 | 12 | 1,059,298 | 74.45 | 1:05:33 | −59.50 |
| **DenseNet-EBCLE** | 16 | 20 | 692,690 | **77.44** | 0:26:33 | |
| ResNeXt-56 | 56 | 16 | 11,004,864 | 58.25 | 3:25:21 | −24.99 |
| **ResNeXt-EBCLE** | 38 | 16 | 7,449,536 | **60.65** | 2:34:09 | |
| **CIFAR-100 dataset** | | | | | | |
| ResNet-50 | 50 | 16 | 766,116 | 40.09 | 0:38:15 | −34.99 |
| **ResNet-EBCLE** | 26 | 24 | 839,428 | **48.57** | 0:24:52 | |
| DenseNet-40 | 40 | 12 | 1,100,428 | **59.27** | 1:07:17 | −62.30 |
| **DenseNet-EBCLE** | 16 | 20 | 725,180 | 58.43 | 0:25:22 | |
| ResNeXt-56 | 56 | 16 | 11,097,024 | 58.89 | 3:43:03 | −29.57 |
| **ResNeXt-EBCLE** | 38 | 16 | 7,541,696 | **60.45** | 2:37:06 | |
| **ImageNet32 dataset** | | | | | | |
| ResNet-50 | 50 | 16 | 824,616 | 33.52 | 14:34:16 | −35.67 |
| **ResNet-EBCLE** | 26 | 24 | 926,728 | **33.57** | 9:22:23 | |
| DenseNet-40 | 40 | 12 | 1,511,728 | **36.72** | 25:43:49 | −33.00 |
| **DenseNet-EBCLE** | 16 | 40 | 1,050,080 | 36.39 | 17:14:25 | |
| ResNeXt-56 | 56 | 16 | 12,018,624 | **36.32** | 85:20:45 | −43.21 |
| **ResNeXt-EBCLE** | 38 | 16 | 8,463,296 | 35.77 | 59:35:47 | |

Table 2. Summary table of results highlighting the relative efficacy of the ResNet models trained adopting the EBCLE heuristic and a dynamic compound scaling approach on the CIFAR-10 benchmarking dataset. EfficientNet-B3-B7 could not be evaluated due to the required memory constraints. ∗ $H'$ = depth and $\chi'$ = breadth co-efficients for EfficientNet models.

| CNN Model | $H'$ | $\chi'$ | Params. | Acc. (%) | Time (h:m:s) |
|---|---|---|---|---|---|
| ResNet-50 | 50 | 16 | 765,098 | 79.85 | 0:38:33 |
| **ResNet-EBCLE** | 26 | 24 | 830,698 | **80.85** | 0:24:34 |
| EfficientNet-B0 | 1.0* | 1.0* | 4,020,358 | 71.29 | 0:25:31 |
| EfficientNet-B1 | 1.1* | 1.0* | 6,525,994 | 75.78 | 1:09:05 |
| EfficientNet-B2 | 1.2* | 1.1* | 7,715,084 | 75.83 | 1:10:25 |

All the models were trained from scratch on the specified hardware utilizing the same source code and libraries. Only the model depth ($H'$) and breadth ($\chi'$), presented in Table 1 had to be modified for baseline comparisons against EBCLE models. In other words, the baseline ResNet-50 model had 50 hidden layers ($H'$) with 16 convolutional units for the first hidden layer ($\chi'$) whereas, EBCLE-models had 26 hidden layers with **24** convolutional units.

## 5. OUR RESULTS

The first set of experiments in this paper is to examine model performance for static CNN architectures with an emphasis on training time with respect to EBCLE-based models, presented as Table 1.

The second set of experiments focused around examining model performance for dynamically scaling CNN architectures such as EfficientNet (EN). The HP configuration used was similar to the settings proposed in [37]; an RMSprop optimizer with the default learning rate of 0.001 and momentum of 0.9 for 100 epochs **with no weight decays or custom layers/objects used to ensure reproducibility**. Furthermore, the image resolution for CIFAR-10 is 32 × 32 but, authors in [12] trained models on the ImageNet dataset with 224 × 224 resolution images, therefore model performance will deteriorate significantly. All models were trained from scratch on the specified hardware.

Finally, we examine the **transfer learning** performance of EBCLE models relative to baseline. The objective is to investigate if limiting depth omits important feature information from being retained that might be pertinent for model performance. The results for transfer learning for STL-10 and CIFAR-10 datasets for the ResNet models were on average **15.75% and 16.93% for the ResNet-50 and ResNet-EBCLE models** respectively. These specific datasets were selected because of their similar constituent class information. All images in the CIFAR-10 dataset were upsampled to normalize pixel resolutions for equalization with the STL-10 dataset. Other models could not be evaluated due to the lack of video memory on the new commissioned hardware.

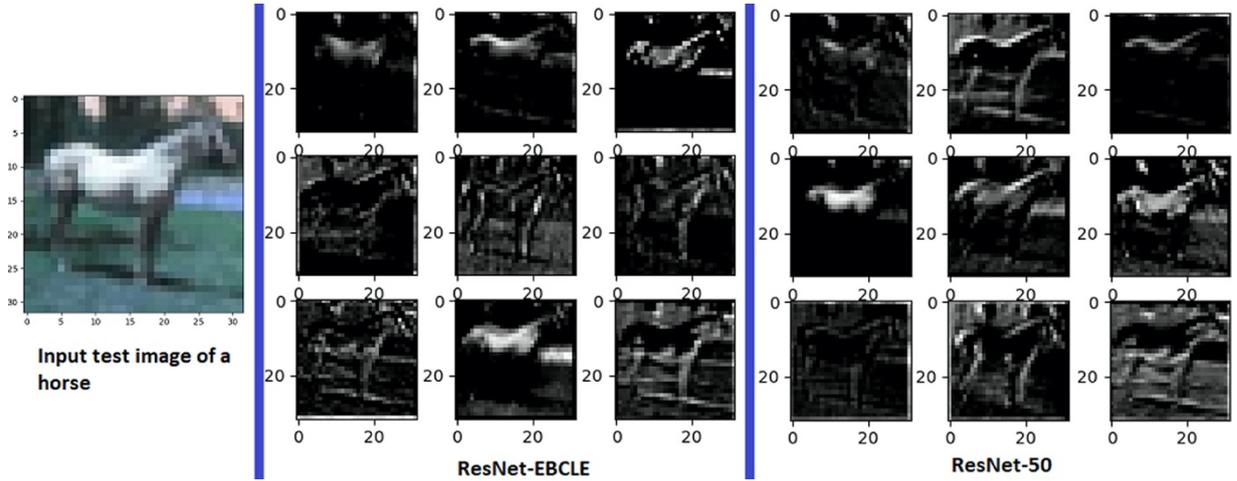

**Fig. 2.** Feature/Activation maps visualized after the first convolutional layer for a test image of a horse in the CIFAR-10 dataset, EBCLE depth = 26.

*5.1. Statistical testing*
First, the Shapiro-Wilk test for normality was used to establish if the collected raw data were normally distributed. The data were normally distributed with all p-values less than 0.05, Table 3 present the mean results over three experimental runs. As the distribution of data is normal, we select the parametric one-tailed paired *t*-test for statistical testing of the data. A one-tailed paired *t*-test is the most applicable since we want to question if there was an observable difference in accuracy and training time for EBCLE models on the same CNN architectures relative to deeper models. In other words, is there a statistical difference in the classification accuracies and training costs when EBCLE models are used instead of the standard CNN models.

Tests were performed with the independent variable as the CNN depth and classification accuracy as the dependent variable. The interpretation was done at the standard significance p-value threshold of 0.05 for a one-tailed test, with the assumption that deeper models should provide higher accuracies when compared to shallower EBCLE models. The default null hypothesis is that no observable differences are present.

## 6. OUR ANALYSIS

In this paper, conventional wisdom advocating the use of deeper CNN models [6] for enhancing classification accuracy has been challenged, with empirical data supporting the validity and efficacy of our proposed novel EBCLE heuristic to significantly reduce model training time without compromising model performance. Examining the input test images in Figs. 2 and 3, the EBCLE models exhibit identical high-level abstractions after the last convolutional layer compared to a deeper ResNet-50 model.

The SE values, measured after the first and last convolutional layer of the ResNet-EBCLE and ResNet-50

**Table 3.** Table of paired one tailed *t*-test results to validating the EBCLE heuristic.

| CNN Models | p-value | mean | variance | (p < 0.05)? |
|---|---|---|---|---|
| **MNIST dataset** | | | | |
| **ResNet-EBCLE** | 0.38 | 99.44 | 0.0086 | No |
| ResNet − 50 | | 99.42 | 0.0003 | |
| **DenseNet-EBCLE** | 0.14 | 99.60 | 0.0064 | No |
| DenseNet − 40 | | 99.54 | 0.0020 | |
| **ResNeXt-EBCLE** | 0.42 | 99.15 | 0.1159 | No |
| ResNext − 56 | | 99.21 | 0.0741 | |
| **CIFAR-10 dataset** | | | | |
| ResNet − 50 | 0.29 | 80.85 | 5.3479 | No |
| **ResNet-EBCLE** | | 79.85 | 0.4723 | |
| DenseNet − 40 | 0.08 | 86.35 | 0.1657 | No |
| **DenseNet-EBCLE** | | 87.22 | 0.1922 | |
| ResNeXt − 56 | 0.32 | 87.29 | 0.8660 | No |
| **ResNext-EBCLE** | | 87.08 | 0.4510 | |
| **STL-10 dataset** | | | | |
| ResNet − 50 | 0.03 | 56.56 | 0.3796 | Yes |
| **ResNet-EBCLE** | | **60.17** | 2.1576 | |
| DenseNet − 40 | 0.04 | 74.45 | 0.1657 | Yes |
| **DenseNet-EBCLE** | | **77.44** | 0.1922 | |
| ResNeXt − 56 | 0.08 | 58.25 | 0.6175 | No |
| **ResNext-EBCLE** | | 60.65 | 1.7851 | |
| **CIFAR-100 dataset** | | | | |
| ResNet − 50 | 0.002 | 40.09 | 0.0409 | Yes |
| **ResNet-EBCLE** | | **48.57** | 1.2637 | |
| DenseNet − 40 | 0.02 | 59.27 | 0.01163 | Yes |
| **DenseNet-EBCLE** | | **62.40** | 0.8481 | |
| ResNext − 56 | 0.081 | 58.89 | 0.0134 | No |
| **ResNeXt-EBCLE** | | 60.45 | 1.8388 | |
| **ImageNet-32 dataset** | | | | |
| ResNet − 50 | 0.44 | 33.52 | 0.3141 | No |
| **ResNet-EBCLE** | | 33.57 | 0.0007 | |
| DenseNet − 40 | 0.21 | 36.72 | 0.1876 | No |
| **DenseNet-EBCLE** | | 36.43 | 0.0030 | |
| ResNext − 56 | 0.16 | 36.32 | 0.3685 | No |
| **ResNeXt-EBCLE** | | 35.77 | 0.4832 | |

models, as visualized in Figs. 2 and 3 are 5.2735 and 5.5625 for the first convolutional layer and 5.3668 and 6.0959 for the last convolutional layer respectively. The difference is more pronounced for the MNIST dataset where the SE measures of the activation maps for the EBCLE and ResNet-50 models after the first convolutional layer are 4.9176 and 4.2341 and after the last convolutional layer are 1.9172 and 2.3010 respectively. The lower SE values in the EBCLE model indicate a higher degree of feature compression in the ResNet-EBCLE model compared to the standard ResNet-50 model with similar higher dimensional feature maps using only half as many convolutional layers thereby maintaining or even outperforming deeper networks.

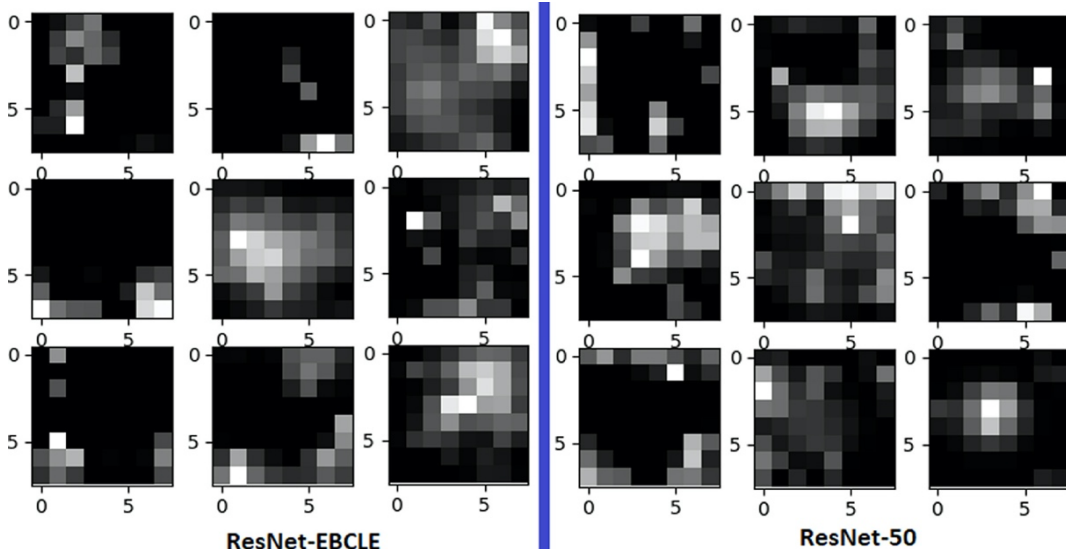

**Fig. 3.** Feature/Activation maps visualized after the last convolutional layer for a test image of a horse in the CIFAR-10 dataset, EBCLE depth = 26.

In a few instances (CIFAR-100 and STL-10), the broader EBCLE models outperformed deeper models by a statistically significant margin, implying that the performance improvements in both accuracies and training times of the EBLCE models are not random. Furthermore, in all instances, the average EBCLE model training time and cost reduction was **45.22%**. The reason is due to the efficient minimization in the trade-off between complexity and information gaps for EBCLE-based models. To discriminate between images related to ships and cars, simple edge detectors that can abstract salient features such as wheels, bow and stern are sufficient to achieve a high classification accuracy. Overly complex abstractions start to increase the information gap while minimizing the complexity gap causing over-fitting and detrimental classification performance.

Utilizing an EBCLE model ensures sufficient dimensionality reduction has occurred before the final classification layer allowing greater fine-grained features to be learned. However, optimality in depth for any CNN model cannot be accurately determined, as asserted by authors in [7]. The proposed EBCLE, at the very least, offers a mathematically sound way to justify HP choices and optimizations affecting classification performance while mitigating untrained features, a characteristic of deep models [9].

A significant contribution of the EBCLE heuristic is the reduction in model training time while maintaining or outperforming baseline classification performance, inline with wider residual network architectures [20]. Other compression, quantization or pruning methods discussed in Section 2.2.1 are accompanied by a statistically significant decrease in classification performance and thus are not studied extensively in this paper.

### 6.1. Exponential increase in trainable parameters leads to marginal gains in performance

The number of trainable parameters increases exponentially for additional convolutional layers, as there is a $2^{\chi'}$ increment in convolutional kernels/units in the model to compensate for the reduction of model capacity [5]. As gradients are in the direction of the steepest descent in back-propagation [11], utilizing unnecessarily deep networks will lead to untrained features [9]. The EBCLE heuristic presented in Section 13 provides an adequate depth estimation using Shannon's entropy for measuring the theoretical limit for feature compression by the convolutional layers after which feature representations resolve into identity transformations which are ineffective in enhancing model performance. Further credence for shallower yet wider models is provided by the data presented in [10], where a classification improvement of 1.1% was achieved from an 117 additional depth increase.

Training CNN models by varying the depths and widths on the same CIFAR-10 dataset while keeping all other HyperParameters constant resulted in the EBCLE model outperforming deeper models. It is noteworthy to mention that additional increases in the initial convolutional width ($\chi'$) caused over-fitting at extremely large values (256) and resulted in decreased classification performance. Moderate values of $\chi'$ (128) produced the best classification accuracy (with an increment of 0.75%) but resulted in an exponential increase in the number of trainable parameters. In other words, models with a $\chi'$ value of 128 had 23,493,130 no. of trainable parameters compared to a narrower model with a $\chi$ value of 24 with only 830,698 i.e. a 96.46% decrease in the number of trainable parameters resulted in only a 0.75% decrease in classification performance. This decrease in the number of parameters suggests excessive model width increases do not offer huge improvements in classification performance similar to very deep models.

#### 6.1.1. Counter-intuitive decrease in model training time

A marked increase in the time required to compute gradients for some EBCLE models due to increase in the total number of parameters can be witnessed relative to baseline models. How- ever, since most of the model training time is consumed during the feature compression phase (discussed in Section 3.1), EBCLE- based models are inherently restricted in terms of their feature compressibility and as such a corresponding decrease in the solution space with yields the observed training time reductions. The increased breadth of the EBCLE-models enables optimal utilization of the computing hardware due to enhanced data loading and parallel processing, relative to deeper networks which experience frequent information processing bottlenecks.

Results presented in Tables 1 and 2 indicate that EBCLE-based models show slight to significant increases in model performance even with decrease in model parameters due to more effective feature extraction, abstraction and compression relative to baseline models. As discussed earlier in Section 6.1, exponential increases in model parameters lead to only marginal gains in performance. As such, more effective training regimes provide significant performance gains compared to simply increasing model sizes. This is due to the tendency of deeper layers to resolve into identity transformations.

*6.2. Limitations*

A key limitation for employing EBCLE is that, the heuristic is limited in applications where the entropic variance of constituent classes in the input dataset is high, as there are no practical ways to determine relative effective variances for individual classes to optimize feature compression. In other words, if some of the constituent classes in the input dataset have high entropy and others have low entropy, EBCLE would not be applicable since mean entropic measures are utilized in this paper.

Another limitation for a comprehensive evaluation of the EBCLE heuristic presented in this paper is that, although the principles of optimizing feature compression should hold true for different application domains or tasks such as audio classification or segmentation; empirical evidence is critical in drawing any meaningful conclusions and as such EBCLE remains confined to CNN image classification in this article.

# 7. CONCLUSION AND FUTURE WORK

To overcome the problems posed by severe over-parameterization concerning model training time and architecture selection, we proposed an entropy-based heuristic that imposes feature abstraction and compression restricting over-parameterization with regards to convolutional depth in CNNs. The proposed heuristic employs a priori knowledge of data distribution for the input dataset to simplify and accelerate CNN model training. Using the EBCLE heuristic, we provided empirical evidence utilizing several well-known benchmarking datasets and CNN architectures against established baselines to validate the efficacy of EBCLE-based models with respect to training time and classification accuracy. Results for the EBCLE heuristic adopting a shallow yet broad CNN model indicate a 24.99% - 78.59% reduction in model training time compared to deeper yet narrower CNN models for the same HyperParameter (HP) configurations without significant performance degradation.

The results presented in this paper support the independent findings obtained in [9], where the authors assert that wider, yet shallower models outperform deeper, yet narrower CNN models. Furthermore, the authors in [38] establish both theoretically and empirically that entropy-based heuristics can simplify and accelerate inner and outer loop calculations for feature selection. Additional validation for our EBCLE heuristic regrading forced feature abstraction and compression can be corroborated by the findings presented in [10], where the authors establish experimentally, that shallower CNN models can learn the same functional representations as deeper networks.

Our proposed EBCLE heuristic offers a simplified approach to select CNN architectures and accelerate model training by utilizing the a priori entropy of input the dataset. Additionally, our EBCLE heuristic is architecturally agnostic facilitating application in multiple domains. The empirical data presented in this paper allude to the same phenomenon of over-parameterization for convolutional widths $\chi'$, suggesting further gains could be achieved with regards to decreasing model training times. This is an important area for exploration and future publications. Empirical validation for our proposed EBCLE heuristic is conducted on five benchmarking datasets (ImageNet32, CIFAR-10/100, STL-10, MNIST) and three network architectures (DenseNet, ResNet, ResNeXt) along with a dynamically scaling network architecture (EfficientNet).

Wider but shallower residual networks have shown to outperform narrow yet deeper networks [20], corroborating the findings presented in this paper. Furthermore, very deep CNN architectures resolve into a collection of independent feature extractors making the process of feature extraction redundant since skip connections facilitate only the most important features to be captured [9]. The EBCLE heuristic could be employed to introduce forced feature abstraction and compression enhancing the efficiency of model training. Empirical evidence supports the fact that shallow residual net- works can learn the same functional representations as deeper networks [10], providing further independent validation that the EBCLE heuristic could help optimize model training, in terms of addressing severe over-parameterization with regards to training time and simplified CNN model selection.

Finally, the theory behind EBCLE for CNN architectures supports the fact that the same principles governing feature compression should apply to other deep learning tasks such as segmentation or regression but, empirical evidence is essential to draw any meaningful conclusions and as such it is reserved as future work.

**CRediT author statement**
Nidhi Gowdra: Conceptualization, Methodology, Software, Validation, Formal analysis, Investigation, Resources, Data Curation, Writing - Original Draft, Writing - Review & Editing, Visualization, Project administration, Funding acquisition. Roopak Sinha: Super- vision, Formal analysis, Writing - Review & Editing. Stephen Mac- Donell: Validation, Formal analysis, Resources, Writing - Review & Editing, Supervision. Wei Qi Yan: Writing - Review & Editing, Supervision.

**Declaration of Competing Interest**

The authors declare the following financial interests/personal relationships which may be considered as potential competing interests:

The corresponding author (Nidhi Gowdra) is the director of the company, InfuseAI Limited (New Zealand) which provided the computational resources for the paper without which this project would have been impossible. The Third author (Stephen MacDonell) is additionally a Professor - Part Time in the information systems and software engineering at the University of Otago, New- Zealand.

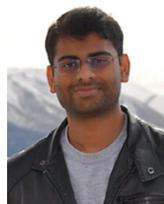

**Nidhi Gowdra** is a third year Ph.D. candidate and a part-time faculty member in the school of engineering and mathematical sciences at Auckland University of Technology (AUT). He is a postgraduate board member as the doctoral representative and the former vice-chair of the IEEE student body at AUT. He is the director of InfuseAI Limited and a certified professional, specialist from numerous fortune 50 companies. He was the chair for Machine and Signal and Image Processing Session at the IEEE Industrial Electronics Society Conference (IECON-2020) and a keynote speaker at the Large Energy Users Forum, New Zealand (2017).

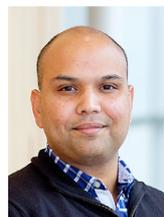

**Roopak Sinha** received his Ph.D. in Electrical and Electronic Engineering from The University of Auckland in 2009. He is currently an Associate Professor at The School of Engineering, Computer and Mathematical Sciences at The Auckland University of Technology, New Zealand. He has previously held academic positions at The University of Auckland, New Zealand and IN-RIA, France. His research interest is Systematic, Standards-First Design of Complex, Next-Generation Embedded Software" applied to domains like Internet-of-Things, Edge Computing,

Cyber-Physical Systems, Home and Industrial Automation, and Intelligent Transportation Systems. He has served on several IEEE/IEC standardisation projects, was an invited editor of the IEEE Transactions on Industrial Informatics, and works with several New Zealand companies to systematically reduce standards-compliance costs in IoT/embedded products.

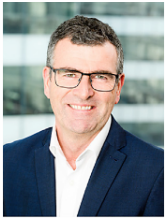

**Stephen MacDonell** is Professor of Software Engineering at Auckland University of Technology and Professor in Information Science at the University of Otago, both in New Zealand. Stephen was awarded BCom(Hons) and MCom degrees from the University of Otago and a Ph.D. from the University of Cambridge. His research has been published in IEEE Transactions on Software Engineering, ACM Transactions on Software Engineering and Methodology, ACM Computing Surveys, Empirical Software Engineering, Information & Management, the Journal of Systems and Software, Information and Software Technology, and the Project Management Journal, and his research findings have been presented at more than 100 international conferences. He is a Fellow of IT Professionals NZ, Senior Member of the IEEE and the IEEE Computer Society, and Member of the ACM, and he serves on the Editorial Board of Information and Software Technology. Stephen is also Theme Leader for Data Science & Digital Technologies in New Zealand's National Science Challenge Science for Technological Innovation, Technical Advisor to the Office of the Federation of Maori Authorities Pou Whakatamore Hangarau - Chief Advisor Innovation & Research, and Deputy Chair of Software Innovation New Zealand.

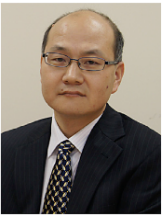

**Wei Qi Yan** is the Director of Centre for Robotics & Vision (CeRV), Auckland University of Technology (AUT); his expertise is in intelligent surveillance and deep learning. Dr. Yan was an exchange computer scientist between the Royal Society of New Zealand (RSNZ) and the Chinese Academy of Sciences (CAS), China. Dr. Yan is a guest (adjunct) professor with PhD supervision of the Chinese Academy of Sciences, China, he was a visiting professor with the National University of Singapore (NUS), the University of Auckland (UOA), and the Massey University, New Zealand.